\documentclass[10pt]{IEEEtran}

\usepackage{caption}
\usepackage{graphicx}
\usepackage{xcolor}
\usepackage{amsmath}
\usepackage{bbold}
\usepackage{booktabs} 
\usepackage[english]{babel}
\usepackage[utf8]{inputenc}
\usepackage[T1]{fontenc}
\usepackage{times}
\usepackage{tabularx}
\usepackage{subfigure}
\usepackage{relsize}
\usepackage{spverbatim}
\usepackage{url}
\usepackage{soul}
\usepackage[subtle]{savetrees}
\usepackage[ruled]{algorithm2e} 
\usepackage{listings}

\newcommand{\tacred}[0]{\textrm{TACRED}}

\title{\textbf{\Large Pynsett: A Programmable Relation Extractor}}

\author{\centering
\small
Alberto Cetoli\\[0.3ex]
QBE Europe\\London, UK\\
alberto.cetoli@uk.qbe.com}

\begin{document}
\maketitle

\begin{abstract}
This paper proposes a programmable relation extraction method for the English language by parsing texts into semantic graphs. 
A person can define rules in plain English that act as matching patterns onto the graph representation. 
These rules are designed to capture the semantic content of the documents, allowing for flexibility and ad-hoc entities. 
Relation extraction is a complex task that typically requires sizable training corpora. 
The method proposed here is ideal for extracting specialized ontologies in a limited collection of documents. 
\end{abstract}


\IEEEpeerreviewmaketitle

\section{Introduction}
The goal of relation extraction is to identify relations among entities in the text. It is an integral part of knowledge base population \cite{Heng2011}, question answering \cite{Xu2016}, and spoken user interfaces \cite{Yoshino:2011:SDS:2132890.2132898}.
Precise relation extraction is still a challenging task
\cite{bunescu-mooney-2005-shortest,Guo2019AttentionGG,Luan2018}, with most existing solutions relying on training data that contains a limited set of relations.

In many useful cases, the relations need to be customized to a specific ontology relevant only in a small collection of documents, making it difficult to acquire enough examples.
This challenge occurs frequently in an industrial context,
where a common solution is string-matching or regular expressions.
Zero-shot learning has been used to overcome this limit: for example, one can understand relation extraction as a question answering problem \cite{Levy2017ZeroShotRE}. This approach can be quite successful, leveraging on recent reading comprehension progress: it trains a system on extracting semantic content first, then applies the learned generalization to create flexible rules for relation extraction. 

While impressive, question answering does not completely solve the challenge of relation extraction,
the major problem being generalizing the query to all the possible variations in which it can be formulated.
Moreover, while using a question answering approach improves the recall of the extractor, it can also lower the precision of the matches due to mistaken reading comprehension.
Representing relations using questions as surface forms does not achieve the same level of precision of rule-based syntactic matches.
For relations of this type, the generalization needed is limited.

Linguistic theories allow to generate a semantic representation that offers a useful generalization of the sentence content, while at the same time providing a framework for precise rule matching.
By using \emph{Discourse Representation Theory} \cite{Kamp1993FromDT} 
or Neo-Davidsonian semantics \cite{Parsons1990}, it is possible to describe a collection of sentences as a set of predicates.
In these frameworks, the relation extraction rules become a pattern matching exercise over graphs. The works of \emph{Reddy et al.} \cite{reddy-etal-2014-large,TACL807} as well as \emph{Tiktinsky et al.} \cite{tiktinsky2020pybart} are an inspiration for this paper.

Further flexibility comes from representing words using word embeddings \cite{MikolovSCCD13}.
In this paper, each lemma is associated to an entry in the \emph{Glove} dataset \cite{pennington2014glove}.
In addition, specialized entities are written as a list of embeddings.

Writing a discourse as a collection of predicates is isomorphic to a graph representation of the text.
The main idea of this paper is to discover relations in the discourse by matching specific sub-graphs. 
Each pattern match is effectively a graph query where the data is the discourse.
The main contribution of this work is two-fold. 
First, it suggests a way to semantically encode sentences.
Second, it defines a method for creating a set of flexible rules for low-resource relation extraction where relations are represented using their surface forms.
The paper is organized as follows: Section \ref{sec:Implementation} describes how the system is implemented. Subsequently, Section \ref{sec:results} discusses some preliminary results, while Section \ref{sec:related_works} summarizes prior works on the subject. Finally, the paper wraps up in Section \ref{sec:conclusions}.
This paper's code can be found at \cite{code-url}.

\section{Implementation}
\label{sec:Implementation}

\subsection{Semantic representation}
\label{sem-repr}
Sentences are transformed into graphs following a similar method to \cite{TACL807}. We start with a dependency parser \cite{spacy2} and apply a series of transformations to obtain a neo-Davidsonian form of the sentence,
where active and passive tenses are represented with the same expression, all words are lemmatized, and co-reference is added to the representation.
For example, the text
\emph{Jane is working at ACME Inc as a woodworker. She is quite taller than the average}
becomes in a predicate form
\begin{lstlisting}{language=Java}
Jane(r1), work(e1), ACME_Inc(r2), woodworker(r3),
AGENT(e1, r1), at(e1, r2), as(e1, r3), 
Jane(r4), be(e2), tall(r5), average(r6), quite(r7),
AGENT(e2, r4), ADJECTIVE(e2, r5), than(r5, r6), 
ADVERB(r5, r7), REFERS_TO(r1, r4), REFERS_TO(r4, r1)
\end{lstlisting}
\begin{figure}[h!]
    \centering
    \includegraphics[width=0.35\textwidth]{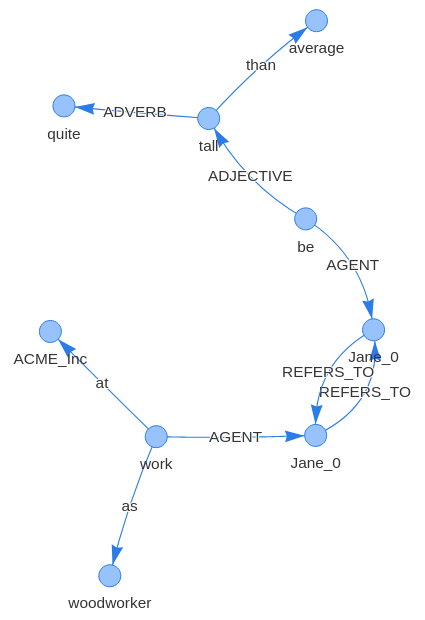}
    \caption{
    \label{fig:graph}
    The text in Section \ref{sem-repr} becomes a semantic graph with co-reference links.}
\end{figure}
In this representation, the text is a graph (Figure \ref{fig:graph}), where the nodes are nouns, verbs, adverbs, and adjectives, and the edges are the semantic relations among them.
The representation used in this work aims to be optimized for the task of extracting relations and for speed.

\subsection{Types of edges}
The main semantic relations employed by the system are explained in the following:

\textbf{AGENT, PATIENT}: the subject and object of the sentence are converted to agent and patient edges coherently with the verb's voice. In addition, these relations are propagated to relevant subordinates.

\textbf{ADJECTIVE, ADVERB}: adjectives and adverbs are connected to the relevant node through an edge. The only exceptions are negation adverbs, which become part of the node's attributes to facilitate the matching procedure, as explained in Section \ref{word-matching}.

\textbf{OWNS}: possessive pronouns are translated into a relation induced by the pronoun's semantics.

\textbf{PREPOSITIONS}: all the prepositions become edges (Figure \ref{fig:graph}). Ideally - in a future work - a further semantic layer should be added to classify the preposition's meaning in context.

\textbf{SUBORDINATES}: the subordinate clauses are linked to the main one through the SUBORDINATE edge. 
One additional type is the ADVOCATIVE\_CLAUSE, marking a conditional relation among sentences. This is a placeholder for future versions of the system where ideally rules can be extracted from the text.

\subsection{Conjunctions}
In order to facilitate graph matching, the conjunction list is flattened and linked to the head node whenever possible. 
For example, the sentence \emph{Jane is smart and wise} becomes, in predicate form
\begin{lstlisting}{language=Java}
Jane(r1), be(e1), smart(r2), wise(r3),
AGENT(e1, r1), ADJECTIVE(e1, r2), ADJECTIVE(e1, r3)
\end{lstlisting}
\begin{figure}[h!]
    \centering
        \subfigure[][]{%
           \includegraphics[width=0.4\textwidth]{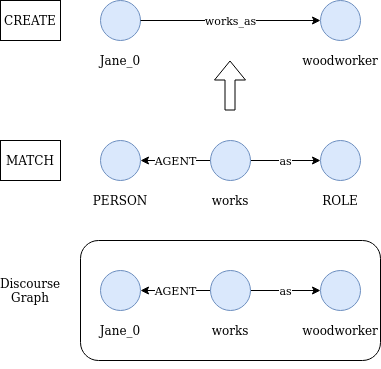}}%
        \qquad
        \subfigure[][]{%
    \includegraphics[width=0.17\textwidth]{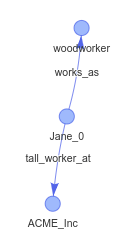}}%
    \qquad \qquad
    \subfigure[][]{%
    \includegraphics[width=0.1\textwidth]{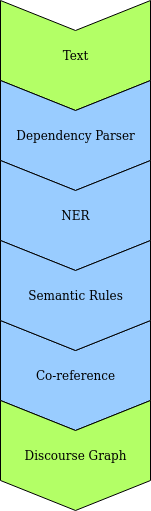}}%
    \caption{
    \label{fig:rules}
    (a) The \emph{MATCH} clause defines the sentence/graph that triggers the rule. 
    The rule then creates an edge between the entities. (b) The resulting relations graph from the two rules in Section \ref{rules}.
    (c) The implementation pipeline transforms an input text onto the discourse graph.}
        \vspace*{-8mm}
\end{figure}
Effectively, 'AND' and 'OR' disappear from the graph. 
This is a crude approximation that facilitates the relation extraction at the expense of semantic correctness.

\subsection{Co-reference}
An additional level of semantics is added by linking together two nouns that co-refer, using the $\mathrm{REFERS\_TO}$ edge.
Currently, the system uses the pre-trained AllenNLP co-reference algorithm \cite{Lee2017EndtoendNC}.
The system - while performing well on the Ontonotes 5 dataset - can increase the noise in the graph by introducing spurious connections.
In order to increase the precision of the model, only pronouns and named entities can match.

\subsection{Matching of words}
\label{word-matching}
Words are represented using the Glove word embeddings of their lemma and a few different tags:
\begin{itemize}
    \item Negated: a \emph{True/False} value that indicates whether a word is associated with a negation: if a verb is negated, the adverb does not appear as a new node, rather the verb is flagged  using this tag. In this way, \emph{work} can never match \emph{does not work}. 
    \item Named Entity Type: a label indicating the entity type of the node, as per \emph{Ontonotes} 5.0 notation \cite{Hovy:2006:O9S:1614049.1614064}.
    \item Node type: indicates whether it is a verb, a noun, an adjective, or an adverb.
\end{itemize}
For example, the noun $\mathrm{Jane}$ is represented internally as
\begin{lstlisting}{language=Java}
{
    vector: EMBEDDING[Jane]
    lemma: "Jane", 
    negated: False, 
    entity_type: PERSON, 
    node_type: noun
}
\end{lstlisting}
Two words match if the dot product between their lemmas' embeddings is greater than a specific threshold, and all the other tags coincide.
For example, the words \emph{carpenter} and \emph{woodworker} match.
This solution can in principle be augmented with an external ontology, where synonyms and hypernyms would trigger a match as well.
In addition, the system allows to cluster a set of words under the same definition.  
\begin{lstlisting}
DEFINE TEAM AS [team, group, club];
DEFINE UNIVERSITY AS [university, academy, polytechnic];
DEFINE LITERATURE AS [book, story, article, series];
\end{lstlisting}
All words within the threshold distance would trigger a match.
For example, the word \emph{tome} would match the word \emph{book}, thus falling into the \emph{LITERATURE} category.

\subsection{Matching of sentences}
\label{rules}

At the end of the processing pipeline, the input text becomes a union of connected graphs, the \emph{discourse graph}. 
The current framework defines rules that act on the discourse graph by declaring two components: a \emph{MATCH} clause, which defines the trigger for the rule, and a \emph{CREATE} clause, which creates the relation edge.
Relations must connect two entities marked by the symbol \textrm{\#}. For example, the sentence \emph{Jane\#1 works at Acme\#2} tags \emph{Jane} and \emph{Acme} for an edge to connect them.
The matching sentence can contain Named Entities (PERSON, ORG, DATE, etc) as well as an internally-defined entity (Section \ref{word-matching}).
An example is as follows
\begin{lstlisting}
DEFINE ROLE AS [carpenter, painter];

MATCH "PERSON#1 works as a ROLE#2."
CREATE (works_as 1 2);

MATCH "PERSON#1 works at ORG#2 as a ROLE. PERSON is tall."
CREATE (tall_worker_at 1 2);
\end{lstlisting}

Please note that a \emph{MATCH} clause is written as a sentence, but it is internally parsed into a graph. A rule is triggered if this semantic representation is a sub-graph of the \emph{discourse graph}. 
Two nodes are considered equal if they match according to the method in Section \ref{word-matching}.
The rules are represented as simple pattern matching rules, as in Figure \ref{fig:rules} (a).
Also, notice that, for the second rule in the above example, more than one sentence is specified. 
This is because the \emph{MATCH} clause can be a text as complex and free-flowing as the documents that are being parsed. The trigger sentences also solve co-reference: in the second rule, the person that works is the same person that is tall.
This second clause expresses the compositional potential of the rules.
In future versions of the framework, one could add more complex mangling of the sentences where simple logical constraints are added (and/or), or information is extracted from mathematical formulas.

Each rule behaves according to the method defined above. When a graph triggers a rule, an edge is created in the \emph{relations graph}, as show in Figure \ref{fig:rules} (b).
In this final representation, the knowledge is condensed into the pre-defined relations.

\subsection{Implementation details}
Every text in the system is processed according to the pipeline in Figure \ref{fig:rules} (c), and eventually transformed into a \emph{discourse graph}.
The dependency parser is Spacy \cite{spacy2}, which also enriches the discourse graph with Named Entities.
Co-reference uses the AllenNLP system, as described in \cite{Lee2017EndtoendNC}.
The semantic transformation rules - available in the open source code 
\cite{rules-url} - 
are implemented through a purpose-made in-memory graph database 
\cite{parvusdb-url}.
Sentences of arbitrary complexity can be parsed by the current system, compatibly with the accuracy of the dependency parser.
This is true both for the input text and the matching rules.

A rule matcher algorithm goes through the list of rules, performs the matching and creates the relations graph according to the method described in Section \ref{rules}.
The computational cost of rule-matching is  $\mathcal{O}(N)$, where $N$ is the number of rules. 
Further improvements should include a rule retriever algorithm, which pre-filters the rules according to the discourse graph. 
Due to speed optimization, the rules are applied only once: the reasoning induced by the rules is only one step deep. 
Ideally, in a future version, the rules should be applied with a Prolog-like resolution tree\cite{kowalski1974}.


\section{Preliminary results}
\label{sec:results}

The current version of the system can be tested against the test set of the \tacred{} corpus \cite{zhang2017tacred}. 
Let us consider only two relations: \emph{date\_of\_birth} and \emph{date\_of\_death} defined as follows

\begin{lstlisting}{language=Prolog}
DEFINE PERSON AS {PERSON};
DEFINE DAY AS {DATE};
DEFINE YEAR AS {DATE};
DEFINE AT_TIME AS {DATE};
DEFINE AT_MOMENT AS [Monday, Tuesday, Wednesday, Thursday, Friday, Saturday, Sunday];

MATCH "PERSON#1 is born AT_TIME#2"
CREATE (DATE_OF_BIRTH 1 2);

MATCH "PERSON#1 is born on DAY#2"
CREATE (DATE_OF_BIRTH 1 2);

MATCH "PERSON#1 is born in YEAR#2"
CREATE (DATE_OF_BIRTH 1 2);

MATCH "PERSON#1 dies AT_MOMENT#2"
CREATE (DATE_OF_DEATH 1 2);

MATCH "PERSON#1 dies on DAY#2"
CREATE (DATE_OF_DEATH 1 2);

MATCH "PERSON#1 dies in YEAR#2"
CREATE (DATE_OF_DEATH 1 2);
\end{lstlisting}

For both relations, there are no false positives (precision is 100\%) while the recall is less competitive: the date of birth relation has 33\% recall, whereas the date of death scores 3.6\%.
This result compares unfavourably with the state of the art on \tacred{} (F1 71.2\% \cite{mtb2019}), 
however, a direct comparison is beyond the scope of this work. The system presented here is not a machine learning model and only aims to create a flexible rule-based framework for precise relation extraction.

\section{Related works}
\label{sec:related_works}
A corpus of works is dedicated to map the output of grammatical parsers onto semantic structures: an early work can be found in the CCGBank manual \cite{ccgbank}, where a set of heuristic rules guide the translation from a constituency parse to a CCG (Categorial Combinatorial Grammar) structure.
Further works \cite{TACL807} apply transformation rules over dependency trees with the goal of achieving logical forms for semantic parsing.
Abstract Meaning Representation \cite{banarescu-etal-2013-abstract} is also used to generate graphs from sentences.

A more recent approach \cite{tiktinsky2020pybart} is tailored to produce enhanced UD Trees (Universal Dependencies Trees) - suited for information extraction tasks - from dependency structures.
The task of extracting relations by using their surface form has been addressed in the influential OpenIE framework \cite{OpenIE2011}.
Similarly, prior work on zero-shot relation extraction \cite{Levy2017ZeroShotRE} attempts to represent relations by using questions.
Hearst patterns \cite{hearst-1992-automatic,roller-etal-2018-hearst} can be used to extract hierarchical relationships from a text, without using semantic representations of documents.
Finally a recent work by \emph{Shlain et al.} \cite{shlain2020syntactic} is closely related to the current paper, where they leverage a syntactic representation of the documents to implement flexible search queries.

\section{Conclusions and future work}
\label{sec:conclusions}
This paper presents a flexible rule-based relation extractor for limited resource sets. 
Flexible rules can be created, thus allowing for a quick relation extractor using specialized ontologies.
The main advantage of this approach is control over the rules and precision in the extracted content.
An extension of the system should allow customized ontologies to be used for word matching.
Moreover, more Named Entities should be included, possibly allowing for specialized extractors within the internal pipeline.
This work uses word embeddings imported from the Glove vectors. A more modern approach could employ pre-trained language models to create the relevant embeddings.
As a final limitation, the system does not assign a temporal dimension to events yet.
This information should be extracted from verb tenses and added to the discourse graph.

\section*{Acknowledgements}
The author is grateful to Stefano Bragaglia for insightful discussions.

\bibliographystyle{IEEEtran}
\bibliography{biblio}

\begin{thebibliography}{10}
\providecommand{\url}[1]{#1}
\csname url@samestyle\endcsname
\providecommand{\newblock}{\relax}
\providecommand{\bibinfo}[2]{#2}
\providecommand{\BIBentrySTDinterwordspacing}{\spaceskip=0pt\relax}
\providecommand{\BIBentryALTinterwordstretchfactor}{4}
\providecommand{\BIBentryALTinterwordspacing}{\spaceskip=\fontdimen2\font plus
\BIBentryALTinterwordstretchfactor\fontdimen3\font minus
  \fontdimen4\font\relax}
\providecommand{\BIBforeignlanguage}[2]{{%
\expandafter\ifx\csname l@#1\endcsname\relax
\typeout{** WARNING: IEEEtran.bst: No hyphenation pattern has been}%
\typeout{** loaded for the language `#1'. Using the pattern for}%
\typeout{** the default language instead.}%
\else
\language=\csname l@#1\endcsname
\fi
#2}}
\providecommand{\BIBdecl}{\relax}
\BIBdecl

\bibitem{Heng2011}
\BIBentryALTinterwordspacing
H.~Ji and R.~Grishman, ``Knowledge base population: Successful approaches and
  challenges,'' in Proceedings of the 49th Annual Meeting of the Association
  for Computational Linguistics: Human Language Technologies - vol. 1, ser. HLT
  '11.\hskip 1em plus 0.5em minus 0.4em\relax Stroudsburg, PA, USA: Association
  for Computational Linguistics, 2011, pp. 1148--1158. [Online]. Available:
  \url{http://dl.acm.org/citation.cfm?id=2002472.2002618}
\BIBentrySTDinterwordspacing

\bibitem{Xu2016}
K.~Xu, S.~Reddy, Y.~Feng, S.~Huang, and D.~Zhao, ``Question answering on
  freebase via relation extraction and textual evidence.''\hskip 1em plus 0.5em
  minus 0.4em\relax Proceedings of the 54th Annual Meeting of the Association
  for Computational Linguistics (vol. 1: Long Papers), 2016, pp. 2326--2336.

\bibitem{Yoshino:2011:SDS:2132890.2132898}
\BIBentryALTinterwordspacing
K.~Yoshino, S.~Mori, and T.~Kawahara, ``Spoken dialogue system based on
  information extraction using similarity of predicate argument structures,''
  in Proceedings of the SIGDIAL 2011 Conference, ser. SIGDIAL '11.\hskip 1em
  plus 0.5em minus 0.4em\relax Stroudsburg, PA, USA: Association for
  Computational Linguistics, 2011, pp. 59--66. [Online]. Available:
  \url{http://dl.acm.org/citation.cfm?id=2132890.2132898}
\BIBentrySTDinterwordspacing

\bibitem{bunescu-mooney-2005-shortest}
\BIBentryALTinterwordspacing
R.~Bunescu and R.~Mooney, ``A shortest path dependency kernel for relation
  extraction,'' in Proceedings of Human Language Technology Conference and
  Conference on Empirical Methods in Natural Language Processing.\hskip 1em
  plus 0.5em minus 0.4em\relax Vancouver, British Columbia, Canada: Association
  for Computational Linguistics, 2005, pp. 724--731. [Online]. Available:
  \url{https://www.aclweb.org/anthology/H05-1091}
\BIBentrySTDinterwordspacing

\bibitem{Guo2019AttentionGG}
Z.~Guo, Y.~Zhang, and W.~Lu, ``Attention guided graph convolutional networks
  for relation extraction,'' in ACL, 2019, p. 241–251.

\bibitem{Luan2018}
\BIBentryALTinterwordspacing
Y.~Luan, L.~He, M.~Ostendorf, and H.~Hajishirzi, ``Multi-task identification of
  entities, relations, and coreference for scientific knowledge graph
  construction,'' in Proceedings of the 2018 Conference on Empirical Methods in
  Natural Language Processing.\hskip 1em plus 0.5em minus 0.4em\relax Brussels,
  Belgium: Association for Computational Linguistics, 2018, pp. 3219--3232.
  [Online]. Available: \url{https://www.aclweb.org/anthology/D18-1360}
\BIBentrySTDinterwordspacing

\bibitem{Levy2017ZeroShotRE}
O.~Levy, M.~Seo, E.~Choi, and L.~S. Zettlemoyer, ``Zero-shot relation
  extraction via reading comprehension,'' in CoNLL.\hskip 1em plus 0.5em minus
  0.4em\relax Proceedings of the 21st Conference on Computational Natural
  Language Learning (CoNLL 2017), 2017, pp. 333--342.

\bibitem{Kamp1993FromDT}
H.~Kamp and U.~Reyle, From Discourse to Logic - Introduction to Modeltheoretic
  Semantics of Natural Language, Formal Logic and Discourse Representation
  Theory.\hskip 1em plus 0.5em minus 0.4em\relax Springer, 1993.

\bibitem{Parsons1990}
T.~Parsons, Events in the Semantics of English.\hskip 1em plus 0.5em minus
  0.4em\relax MIT Press, 1990.

\bibitem{reddy-etal-2014-large}
S.~Reddy, M.~Lapata, and M.~Steedman, ``Large-scale semantic parsing without
  question-answer pairs,'' Transactions of the Association for Computational
  Linguistics, vol.~2, 2014, pp. 377--392.

\bibitem{TACL807}
S.~Reddy, O.~Tackstrom, M.~Collins, T.~Kwiatkowski, D.~Das, M.~Steedman, and
  M.~Lapata, ``Transforming dependency structures to logical forms for semantic
  parsing,'' Transactions of the Association for Computational Linguistics,
  vol.~4, 2016, pp. 127--140.

\bibitem{tiktinsky2020pybart}
\BIBentryALTinterwordspacing
A.~Tiktinsky, Y.~Goldberg, and R.~Tsarfaty, ``pybart: Evidence-based syntactic
  transformations for {IE},'' in Proceedings of the 58th Annual Meeting of the
  Association for Computational Linguistics: System Demonstrations, {ACL} 2020,
  Online, July 5-10, 2020, A.~{\c{C}}elikyilmaz and T.~Wen, Eds.\hskip 1em plus
  0.5em minus 0.4em\relax Association for Computational Linguistics, 2020, pp.
  47--55. [Online]. Available:
  \url{https://www.aclweb.org/anthology/2020.acl-demos.7/}
\BIBentrySTDinterwordspacing

\bibitem{MikolovSCCD13}
\BIBentryALTinterwordspacing
T.~Mikolov, I.~Sutskever, K.~Chen, G.~Corrado, and J.~Dean, ``Distributed
  representations of words and phrases and their compositionality,'' CoRR, vol.
  abs/1310.4546, 2013. [Online]. Available:
  \url{http://arxiv.org/abs/1310.4546}
\BIBentrySTDinterwordspacing

\bibitem{pennington2014glove}
\BIBentryALTinterwordspacing
J.~Pennington, R.~Socher, and C.~D. Manning, ``Glove: Global vectors for word
  representation,'' in Empirical Methods in Natural Language Processing
  (EMNLP), 2014, pp. 1532--1543. [Online]. Available:
  \url{http://www.aclweb.org/anthology/D14-1162}
\BIBentrySTDinterwordspacing

\bibitem{code-url}
\BIBentryALTinterwordspacing
``Pynsett source code,'' 2020. [Online]. Available:
  \url{https://github.com/fractalego/pynsett/tree/semapro2020}
\BIBentrySTDinterwordspacing

\bibitem{spacy2}
M.~Honnibal and I.~Montani, ``{spaCy 2}: Natural language understanding with
  {B}loom embeddings, convolutional neural networks and incremental
  parsing.''\hskip 1em plus 0.5em minus 0.4em\relax ExplosionAI, 2017.

\bibitem{Lee2017EndtoendNC}
K.~Lee, L.~He, M.~Lewis, and L.~Zettlemoyer, ``End-to-end neural coreference
  resolution,'' in EMNLP, 2017, pp. 188--197.

\bibitem{Hovy:2006:O9S:1614049.1614064}
\BIBentryALTinterwordspacing
E.~Hovy, M.~Marcus, M.~Palmer, L.~Ramshaw, and R.~Weischedel, ``Ontonotes: The
  90
  the NAACL, Companion Volume: Short Papers, ser. NAACL-Short '06.\hskip 1em
  plus 0.5em minus 0.4em\relax Stroudsburg, PA, USA: Association for
  Computational Linguistics, 2006, pp. 57--60. [Online]. Available:
  \url{http://dl.acm.org/citation.cfm?id=1614049.1614064}
\BIBentrySTDinterwordspacing

\bibitem{rules-url}
\BIBentryALTinterwordspacing
``List of transformation rules,'' 2020. [Online]. Available:
  \url{https://github.com/fractalego/pynsett/tree/semapro2020/pynsett/rules/parsing}
\BIBentrySTDinterwordspacing

\bibitem{parvusdb-url}
\BIBentryALTinterwordspacing
``In-memory graph database,'' 2020. [Online]. Available:
  \url{https://github.com/fractalego/parvusdb}
\BIBentrySTDinterwordspacing

\bibitem{kowalski1974}
R.~Kowalski, ``Predicate logic as programming language,'' vol.~74.\hskip 1em
  plus 0.5em minus 0.4em\relax IFIP Congr., 1974, pp. 569--574.

\bibitem{zhang2017tacred}
\BIBentryALTinterwordspacing
Y.~Zhang, V.~Zhong, D.~Chen, G.~Angeli, and C.~D. Manning, ``Position-aware
  attention and supervised data improve slot filling,'' in Proceedings of the
  2017 Conference on Empirical Methods in Natural Language Processing (EMNLP
  2017), 2017, pp. 35--45. [Online]. Available:
  \url{https://nlp.stanford.edu/pubs/zhang2017tacred.pdf}
\BIBentrySTDinterwordspacing

\bibitem{mtb2019}
L.~Baldini~Soares, N.~FitzGerald, J.~Ling, and T.~Kwiatkowski, ``Matching the
  blanks: Distributional similarity for relation learning.''\hskip 1em plus
  0.5em minus 0.4em\relax Proceedings of the 57th Annual Meeting of the
  Association for Computational Linguistics, 2019, pp. 2895--2905.

\bibitem{ccgbank}
\BIBentryALTinterwordspacing
J.~Hockenmaier and M.~Steedman, ``Ccgbank: User's manual,'' 2005. [Online].
  Available:
  \url{https://catalog.ldc.upenn.edu/docs/LDC2005T13/CCGbankManual.pdf}
\BIBentrySTDinterwordspacing

\bibitem{banarescu-etal-2013-abstract}
\BIBentryALTinterwordspacing
{L. Banarescu et al.}, ``{A}bstract {M}eaning {R}epresentation for
  sembanking,'' in Proceedings of the 7th Linguistic Annotation Workshop and
  Interoperability with Discourse.\hskip 1em plus 0.5em minus 0.4em\relax
  Sofia, Bulgaria: Association for Computational Linguistics, 2013, pp.
  178--186. [Online]. Available:
  \url{https://www.aclweb.org/anthology/W13-2322}
\BIBentrySTDinterwordspacing

\bibitem{OpenIE2011}
\BIBentryALTinterwordspacing
J.~Christensen, Mausam, S.~Soderland, and O.~Etzioni, ``An analysis of open
  information extraction based on semantic role labeling,'' in Proceedings of
  the Sixth International Conference on Knowledge Capture, ser. K-CAP
  ’11.\hskip 1em plus 0.5em minus 0.4em\relax New York, NY, USA: Association
  for Computing Machinery, 2011, p. 113–120. [Online]. Available:
  \url{https://doi.org/10.1145/1999676.1999697}
\BIBentrySTDinterwordspacing

\bibitem{hearst-1992-automatic}
\BIBentryALTinterwordspacing
M.~A. Hearst, ``Automatic acquisition of hyponyms from large text corpora,'' in
  {COLING} 1992 vol. 2: The 15th {I}nternational {C}onference on
  {C}omputational {L}inguistics, 1992. [Online]. Available:
  \url{https://www.aclweb.org/anthology/C92-2082}
\BIBentrySTDinterwordspacing

\bibitem{roller-etal-2018-hearst}
\BIBentryALTinterwordspacing
S.~Roller, D.~Kiela, and M.~Nickel, ``Hearst patterns revisited: Automatic
  hypernym detection from large text corpora,'' in Proceedings of the 56th
  Annual Meeting of the Association for Computational Linguistics (vol 2: Short
  Papers).\hskip 1em plus 0.5em minus 0.4em\relax Melbourne, Australia:
  Association for Computational Linguistics, 2018, pp. 358--363. [Online].
  Available: \url{https://www.aclweb.org/anthology/P18-2057}
\BIBentrySTDinterwordspacing

\bibitem{shlain2020syntactic}
\BIBentryALTinterwordspacing
M.~Shlain, H.~Taub-Tabib, S.~Sadde, and Y.~Goldberg, ``Syntactic search by
  example,'' 2020. [Online]. Available: \url{https://arxiv.org/abs/2006.03010}
\BIBentrySTDinterwordspacing

\end{thebibliography}
\end{document}